\documentclass{IOS-Book-Article}

\usepackage{mathptmx}

\usepackage{soul}\setuldepth{article}
\usepackage{url}
\usepackage{booktabs}
\usepackage{xcolor}
\def\hb{\hbox to 10.7 cm{}}

\begin{document}

\pagestyle{headings}
\def\thepage{}

\begin{frontmatter}              

\title{Evaluating multilingual BERT for Estonian}


\author{\fnms{Claudia} \snm{Kittask}%
\thanks{Corresponding Author: Claudia Kittask; E-mail: claudiakittask@gmail.com
}}
and
\author{\fnms{Kirill} \snm{Milintsevich}}
and
\author{\fnms{Kairit} \snm{Sirts}}

\address{Institute of Computer Science, University of Tartu, Tartu, Estonia}

\begin{abstract}
Recently, large pre-trained language models, such as BERT, have reached state-of-the-art performance in many natural language processing tasks, but for many languages, including Estonian, BERT models are not yet available. However, there exist several multilingual BERT models that can handle multiple languages simultaneously and that have been trained also on Estonian data. In this paper, we evaluate four multilingual models---multilingual BERT, multilingual distilled BERT, XLM and XLM-RoBERTa---on several NLP tasks including POS and morphological tagging, NER and text classification. Our aim is to establish a comparison between these multilingual BERT models and the existing baseline neural models for these tasks.
Our results show that multilingual BERT models can generalise well on different Estonian NLP tasks outperforming all baselines models for POS and morphological tagging, text classification and NER,
with XLM-RoBERTa achieving the highest results compared with other multilingual models. 
\end{abstract} 
\end{frontmatter}

\section{Introduction}

Large pretrained language models, also called contextual word embeddings, such as ELMo \cite{peters2018deep} or BERT \cite{devlin2019bert} have been shown to improve many natural language processing tasks.
Training large contextual language models is complex both in terms of the required computational resources as well as the training process and thus, the number of languages for which the pretrained models are available is still limited. 

Although, according to \cite{nozza2020mask}, language-specific BERT models are currently available for 19 languages, many more languages are supported via multi-lingual models. The aim of the multilingual models is to reduce the necessity to train language-specific models for each language separately. Experiments on various tasks, such as named entity recognition (NER) \cite{conneau-etal-2020-unsupervised} or parsing pipeline tasks \cite{kondratyuk201975}, have shown that multilingual contextual models can help to improve the performance over the baseline models not based on contextual word embeddings. 

There are several multilingual models available that also include Estonian language. For instance, multilingual BERT (mBERT) \cite{devlin2019bert} has been trained jointly on Wikipedia data on 104 languages, including also Estonian.
Estonian is also included in the cross-lingual language model (XLM-100) \cite{NIPS2019_8928}, which was trained on 100 Wikipedia languages, and cross-lingual RoBERTa (XLM-RoBERTa) \cite{conneau-etal-2020-unsupervised}, which was trained on much larger CommonCrawl corpora and also includes 100 languages.
Finally, DistilBERT \cite{sanh2019distilbert} is a smaller version of the BERT model obtained from the BERT models via knowledge distillation, which is a compression technique where the compact model is trained to reproduce the behaviour of the larger model.
The multilingual DistilBERT (DistilmBERT) has been distilled from the mBERT model featuring the same 104 Wikipedia languages.

The aim of the current work is to evaluate the existing multilingual BERT models on several NLP tasks on Estonian. In particular, we will apply the BERT models on NER, POS and morphological tagging, and text classification tasks.  We compare four multilingual models---mBERT, XLM-100, XLM-RoBERTa and DistilmBERT---to find out which one of those performs the best on our Estonian tasks.  
We compare the results of the multilingual BERT models with task-specific baselines and show that multilingual BERT models improve the performance of the Estonian POS and morphological tagging, text classification tasks and named entity recognition. 
Overall, XLM-RoBERTa achieves the best results compared with other multilingual BERT models used.  


\section{Related work}

Although most research on multilingual BERT models has been concerned about zero-shot cross-lingual transfer \cite{pires2019multilingual}, we are more interested in those previous works that, similar to us, evaluate multilingual BERT models in comparison to monolingual (non-English) baselines. We next review some examples of such work.

Virtanen et al. \cite{virtanen2019multilingual} evaluated multilingual BERT alongside with the monolingual Finnish BERT on several NLP tasks. In their work, multilingual BERT models outperformed monolingual baselines for text classification and NER tasks, while for POS-tagging and dependency parsing the multilingual BERT models fell behind the previously proposed methods, most of which were utilizing monolingual contextual ELMo embeddings \cite{peters2018deep}.
Baumann \cite{baumann2019multilingual} evaluated multilingual BERT models on German NER task and found that while the multilingual BERT models outperformed two non-contextual LSTM-CRF-based baselines, it performed worse than a model utilizing monolingual contextual character-based string embeddings \cite{akbik2018contextual}.
Kuratov et al. \cite{kuratov2019adaptation} applied multilingual BERT models on several tasks in Russian. They found that multilingual BERT outperformed non-contextual baselines for paraphrase identification and question answering and fell below a baseline for sentiment classification.

The pattern in all these works is similar: the multilingual BERT models perform better than non-neural or non-contextual neural baselines but the multilingual BERT model is typically outperformed by approaches based on language-specific monolingual contextual comparison systems. We cannot test the second part of this observation as currently no monolingual language-specific BERT model exist for Estonian. However, we will show that the first part of this observation generally holds also for Estonian, i.e. the multilingual BERT models outperform non-contextual baselines for most of the experimental tasks used in this paper.

\section{Experimental Tasks}
\label{sec:tasks}

This section describes the experimental tasks. We give also overview of the used data and the baseline models. 

\subsection{POS and Morphological Tagging}

For POS and morphological tagging, we use the Estonian treebank from the Universal Dependencies (UD) v2.5 collection that contains annotations of lemmas, part of speech, universal morphological features, dependency heads and universal dependency labels. We train models to predict both universal POS (UPOS) and language-specific POS (XPOS) tags as well as morphological tags. 
We use the pre-defined train/dev/test splits for training and evaluation. Table~\ref{tab:ud_stats} shows the statistics about the treebank splits.

\begin{table}[h]
\begin{tabular}{llll}
\toprule
\textbf{} & \textbf{Train} & \textbf{Dev} & \textbf{Test} \\
\midrule
Sentences & 31012          & 3128         & 6348          \\
Tokens  & 344646         & 42722        & 48491 \\
\bottomrule
\end{tabular}
\caption{\label{tab:ud_stats}Statistics for the Estonian UD corpus.}
\end{table}

As baselines, we report the results of Stanza \cite{qi2020stanza} and UDPipe \cite{straka2017tokenizing} obtained on the same Estonian UD v2.5 test set.

\subsection{Article Type and Sentiment Classification}

\begin{table}[b]
\centering
\begin{tabular}{lcccc|c}
\toprule

                 & \multicolumn{1}{c}{ \bf Negative} & \multicolumn{1}{c}{ \bf Ambiguous} & \multicolumn{1}{c}{ \bf Positive} & \multicolumn{1}{c}{ \bf Neutral} & \multicolumn{1}{|c}{ \bf Total} \\
\midrule
\bf Opinion          & 429                          & 242                           & 162                          & 139                         & 972                       \\
\bf Estonia          & 152                          & 41                            & 93                           & 133                         & 419                       \\
\bf Life             & 138                          & 47                            & 207                          & 128                         & 520                       \\
\bf Comments-Life    & 347                          & 40                            & 79                           & 41                          & 507                       \\
\bf Comments-Estonia & 368                          & 27                            & 50                           & 56                          & 501                       \\
\bf Crime            & 170                          & 12                            & 11                           & 16                          & 209                       \\
\bf Culture          & 57                           & 40                            & 86                           & 79                          & 262                       \\
\bf Sports            & 76                           & 81                            & 152                          & 76                          & 385                       \\
\bf Abroad           & 190                          & 22                            & 42                           & 59                          & 313                       \\
\midrule
\bf Total            & 1927                         & 552                           & 882                          & 727                         & 4088  \\
\bottomrule
\end{tabular}
\caption{\label{tab:text_stats}Statistics of the Estonian Valence corpus.}
\end{table}

For text classification, we use the Estonian Valence corpus \cite{Pajupuu2016IDENTIFYINGPI}, which consists of 4088 paragraphs obtained from Postimees daily. The corpus has been annotated with sentiment as well as with rubric labels. The statistics of this dataset are given in Table~\ref{tab:text_stats}.
We split the data into training, testing and development set using 70/20/10 split preserving the ratios of different labels in the splits. 
All duplicates were removed from the corpus. In total, there were 17 duplicate paragraphs. We followed the suit of Pajupuu et al. \cite{Pajupuu2016IDENTIFYINGPI} and removed the paragraphs annotated as ambiguous from the corpus. These paragraphs were shown to considerably lower the accuracy of the classification.   

For baseline, we trained supervised fastText classifiers \cite{joulin2017bag} with pretrained fastText Wiki embeddings. The best hyperparameter values were found using the built-in fastText hyperparameter optimization.

\subsection{Named Entity Recognition}

The available Estonian NER corpus was created by Tkachenko et al. \cite{tkachenko2013named}. The corpus annotations cover three types of named entities: locations, organizations and persons. It contains 572 news stories published in local online newspapers Postimees and Delfi covering local and international news on a range of different topics. We split the data into training, testing and development set using 80/10/10 splits while preserving the document boundaries. Table~\ref{tab:ner_stats} shows statistics of the splits. 

\begin{table}[h!]
\begin{tabular}{cccccc|c}
\toprule
      & \bf Sentences & \bf Tokens & \bf{PER} & \bf{LOC} & \bf{ORG} & \bf{Total} \\
      \midrule
Train & 9965 & 155981 & 6174         & 4749         & 4784         & 15707          \\
Dev   & 2429 & 32890 & 1115          & 918          & 742          & 2775           \\
Test  & 1908 & 28370 & 1201         & 644          & 619          & 2464 \\
\bottomrule
\end{tabular}
\caption{\label{tab:ner_stats}Statistics of the Estonian NER corpus.}
\end{table}

As baselines, we report the performance of the CRF model \cite{tkachenko2013named} and the bilinear LSTM sequence tagger that was adapted from the Stanza POS tagger \cite{qi2020stanza}. The tagger was trained on the NER annotations instead of POS tags, and the input was enriched with both POS tags and morphological features, i.e. the input to the NER model was the concatenation of the word, and its POS and morphological tag embeddings. The POS and morphological tags were predicted with the pre-trained Stanza POS tagger.
The entity level performance is evaluated using the conlleval script from CoNLL-2000 shared task.

\section{Experimental setup}

We conduct experiments with four different multilingual BERT models: multilingual cased BERT-base (mBERT), multilingual cased DistilBERT (DistilmBERT), cased XLM-100 and cross-lingual RoBERTa (XLM-RoBERTa).
All these models are available via Hugging Face transformers library\footnote{\url{https://huggingface.co/transformers/}}.
Each model is available with sequence lengths of 128 and 512 and we experiment with both.  Table~\ref{tab:stats} shows some details of the models. 

\begin{table}[ht!]
\begin{tabular}{lccc}
\toprule
            & \bf Languages  & \bf Vocab size & \bf Parameters \\
\midrule
mBERT       & 104       &     119K    & 172M            \\
XLM-100     & 100        &  200K      & 570M            \\
DistilmBERT & 100         &   119K    & 66M     \\
XLM-RoBERTa & 100 & 250K & 270M \\
\bottomrule
\end{tabular}
\caption{\label{tab:stats}Details of multilingual BERT models (all cased)}
\end{table}

To evaluate the performance of the multilingual BERT models on downstream tasks, we fine-tune all four BERT models for the NLP tasks described in Section~\ref{sec:tasks}. In addition to training the task-specific classification layer, we also fine-tune all BERT model parameters as well. 
For data processing and training, we used the scripts publicly available in the Hugging Face transformers repository.
We tune the learning rate of the AdamW optimizer and batch size for each multilingual model and task on the development set using grid search. The learning rate was searched from the set of (5e-5, 3e-5, 1e-5, 5e-6, 3e-6). Batch size was chosen from the set of (8, 16). 
We find the best model for each learning rate and batch size combination by using early stopping with patience of 10 epochs on the development set.

\section{Results}

In subsequent sections we present the experimental results on all multilingual BERT models for POS and morphological tagging, text classification and named entity recognition tasks.

\subsection{POS and morphological tagging}

%

\begin{table}[b!]
\begin{tabular}{lcccccc}
\toprule
 \bf Model & \bf UPOS  & \bf XPOS & \bf Morph & \bf UPOS  & \bf XPOS & \bf Morph \\
 & \multicolumn{3}{c}{\bf Seq = 128} & \multicolumn{3}{c}{\bf Seq = 512} \\
\midrule
mBERT                  & 97.42 & 98.06 & 96.24 & 97.43 & 98.13 & 96.13  \\
DistilmBERT            & 97.22 & 97.75 & 95.40 & 97.12 & 97.78 & 95.63  \\
XLM-100                & 97.60 & 98.19 & \bf 96.57 & 97.59 & 98.06 & 96.54          \\
XLM-RoBERTa            &\bf 97.78 &\bf 98.36 & 96.53 & \bf97.80 & \bf98.40 & \bf 96.69    \\
\midrule
Stanza \cite{qi2020stanza}  & 97.19 & 98.04 & 95.77 \\
UDPipe \cite{straka2017tokenizing}  & 95.7 & 96.8 & 93.5  \\
\bottomrule
\end{tabular}
\caption{\label{tab:pos_results}POS and morphological tagging accuracy on Estonian UD test set.}
\end{table}

The results for POS and morphological tagging are summarized in Table~\ref{tab:pos_results}.
In general, all tested multilingual BERT models are equally good and perform better than the Stanza and UDPipe baselines. DistilmBERT was the only multilingual model that did not exceed the baseline models results. On the other hand, the XLM-RoBERTa stands out with a small but consistent improvement over all other results displayed.  Results also show that the sequence length of the model does not affect the performance in any way. The performance on XPOS is better than on UPOS. This is probably caused by the difference in the POS tag annotation schemes. 

\subsection{Text classification}

\begin{table}[t]
\begin{tabular}{lccll}
\toprule
\textbf{Model} & \textbf{Rubric}      & \textbf{Sentiment}   & \multicolumn{1}{c}{\textbf{Rubric}} & \textbf{Sentiment} \\
               & \multicolumn{2}{c}{\textbf{Seq = 128}}      & \multicolumn{2}{c}{\textbf{Seq = 512}} \\
\midrule
mBERT       &  75.67        & 70.23   & 74.94  &  69.52  \\
DistilmBERT &   74.57           &  65.95  & 74.93 & 66.95   \\
XLM-100     &    76.78          &   73.50 & 77.15 & 71.51           \\
XLM-RoBERTa &   \bf 80.34     &   \bf 74.50  & \bf  78.62   & \bf 76.07          \\
\midrule
fastText    &   71.01      &  66.76             \\
\bottomrule
\end{tabular}
\caption{\label{tab:sent_results} Rubric and sentiment classification accuracy.}
\end{table}

The sentiment and rubric classification task results are shown in Table~\ref{tab:sent_results}.
Multilingual models can easily outperform baseline fastText model. Similarly to POS and morphological tagging tasks, XLM-RoBERTa achieved the highest and DistilmBERT the lowest results overall. Even though there are more labels the in rubric classification task, it is still easier for the models to correctly classify than the sentiment classification task. Comparison between the models with different sequence lengths is inconclusive---in some cases the models with longer sequence are better but not always.

\subsection{Named Entity Recognition}




\begin{table}[b]
\begin{tabular}{lcccccc|ccc}
\toprule
 \bf Model & \bf Prec  & \bf Recall & \bf F1 & \bf Prec  & \bf Recall & \bf F1 & \bf Prec  & \bf Recall & \bf F1  \\
 & \multicolumn{3}{c}{\bf Seq = 128} & \multicolumn{3}{c|}{\bf Seq = 512} & \multicolumn{3}{c}{\bf Seq = Concatenated}\\
\midrule
mBERT                  & 85.88 & 87.09 & 86.51 & \bf 88.47 & 88.28 & 88.37 & 86.42    & 89.64    & 88.01 \\
DistilmBERT            & 84.03 & 86.98 & 85.48 & 85.30 & 86.49 & 85.89 & 83.18    & 87.38    & 85.23 \\
XLM-100                & \bf 88.16 & 88.11 & 88.14 & 87.86 & 89.52 & 88.68 & 73.27    & 80.48    & 76.71\\ 
XLM-RoBERTa            & 87.55 & \bf 91.19 & \bf 89.34 & 87.50 & \bf 90.76 & \bf \bf 89.10 & \bf 87.69    & \bf 92.70    & \bf 90.12\\
\midrule

CRF & 84.41 & 85.09 & 84.75  \\
StanfordNLP  & 88.33 & 90.38 & 89.35  \\
\bottomrule
\end{tabular}
\caption{\label{tab:ner_results}NER tagging results. The right-hand part of the table shows the results with the models of sequence length 512, with the input sentences concatenated into sequences of maximum length.}
\end{table}

The Table~\ref{tab:ner_results} (left) summarizes the NER results. The multilingual XLM-RoBERTa is superior to all multilingual models and better or as good as task-specific models. In particular, the XLM-RoBERTa results are comparable with the task-specific StanfordNLP model, which achieved the best results after XLM-RoBERTa. 
CRF based model was easily outperformed by all multilingual models except for DistilmBERT.

While performing these experiments, each sentence was treated as one sequence.
This may have not optimally used the maximum sequence length available, especially in models with sequence length 512. 
As most sentences in our NER corpus do not reach the maximum length, we hypothesize that
using longer sequences with the models of sequence length 512 would add more context for the model and thus improve the results.
For that, we concatenate sentences from the same document to reach to the maximum 512 wordpiece sequence. The right-most section of the Table \ref{tab:ner_results} shows the results of the experiments with longer input sequences.
The numbers in the table show that concatenating the input sequences does not boost the scores. Compared with the regular results based on single sentences, only XLM-RoBERTa was able to utilize the maximum sequence length while the scores of other models decreased. The performance of the XLM-100 model suffered the most and obtained even lower results than DistilmBERT, which so far has gotten the lowest results in all tasks. 

One possible reason why the multilingual BERT models were not able to improve over the Stanford tagger based NER model is that the Stanford baseline model makes use of the POS and morphological information while the BERT models do not.
Adding POS and/or morphological information the BERT model has the potential to improve their results, as especially POS information can be crucial for detecting proper names that make up a large number of named entities.
 
We experimented with two different approaches for adding POS and morphological information to the BERT-based models.
The first approach (\textsc{Pre-BERT}) only changes the input of the models. Here, the POS and morphological information is input directly into the BERT model by adding the embeddings of POS and morphological tags to the default input embeddings by summing all embedding vectors. 
The second approach (\textsc{Post-BERT}) requires slight changes in the sequence classification model. Here, the embeddings of POS and morphological tags are concatenated to the output vector obtained from the BERT model and the concatenated representation is then input to the classification layer.
We expect the \textsc{Post-BERT} method to perform better because in this approach, the POS and morphological information is fed to the model closer to the classification layer and thus has the more direct influence on the classification decision. The advantage of the \textsc{Pre-BERT} approach, on the other hand, is its simplicity as it does not require any changes in the model architecture.
For training with both approaches we used the POS and morphological information supplied with the NER corpus. The POS and morphological tags for the test part were obtained with the open-source Estonian morphological analyzer Vabamorf \cite{kaalep1997estonian} that uses the same annotation scheme as supplied in the NER corpus. 


\begin{table}[t]
\centering
\begin{tabular}{lccccccc}
\toprule
            & \multicolumn{3}{c}{\textsc{Pre-BERT}} & \multicolumn{3}{c}{\textsc{Post-BERT}} & Regular \\
            & POS+Morph   & POS     & Morph  & POS+Morph   & POS     & Morph  & -       \\
\midrule
mBERT       & 82.58       & 83.80   & 86.41 & 87.10       & \bf 88.59   & 87.13    & 86.51   \\
distilmBERT & 70.30       & 79.39   & 82.16  & 81.84       & 83.51   & 84.97    & \bf 85.48   \\
XLM-100     & 80.26       & 82.48   & 87.36 & 81.25       & 86.76   & 86.42    & \bf 88.14   \\
XLM-RoBERTa &  89.71       & \bf 89.86   & 89.43    & 89.52       & 86.76   & 87.62    & 89.34  \\
\bottomrule
\end{tabular}
\caption{\label{tab:extra_results} NER F1 scores with additional POS and morphological information. }
\end{table}

Table~\ref{tab:extra_results} shows that the results of adding POS and/or morphological features is mixed. While mBERT achieves a large improvement and XLM-RoBERTa a marginal increase in performance, the scores of other two models actually decrease quite a bit.
Overall, as expected, the \textsc{Post-BERT} approach, where the extra features are concatenated to the output vector of BERT, is better than the \textsc{Pre-BERT} approach. The exception is again the XLM-RoBERTa model that with the \textsc{Pre-BERT} method achieves the best NER results of all multilingual models and now also improves over the Stanford tagger based baseline.
From the three settings
adding only POS or morphological features seems the best, except again for XML-RoBERTa, for which also the combination of POS and morphological information seems beneficial.
To conclude, adding either POS and/or morphological features can be helpful for the mBERT and XLM-RoBERTa models, other two models were not able to use the extra features to increase the scores.

\section{Conclusions}

In this work, we compared multilingual BERT and BERT-like models with non-contextual baseline models on several downstream NLP tasks. For all tasks,
multilingual models outperformed the previously proposed task-specific models, with XLM-RoBERTa achieving the highest scores on all experimental tasks, while DistilmBERT performed the worst overall. Based on these results we can recommend using the XLM-RoBERTa as a basis for neural NLP models for Estonian.  
Considering the results from previous works comparing multilingual BERT with language-specific BERT models \cite{nozza2020mask,virtanen2019multilingual}, further performance gains can be obtained from training monolingual BERT for Estonian, in particular following the RoBERTa guidelines \cite{liu2019roberta}.

\bibliographystyle{plain}
\bibliography{refs}

\begin{thebibliography}{10}

\bibitem{akbik2018contextual}
Alan Akbik, Duncan Blythe, and Roland Vollgraf.
\newblock {Contextual String Embeddings for Sequence Labeling}.
\newblock In {\em Proceedings of COLING}, pages 1638--1649, 2018.

\bibitem{baumann2019multilingual}
Antonia Baumann.
\newblock {Multilingual Language Models for Named Entity Recognition in German
  and English}.
\newblock In {\em Proceedings of RANLP SRW 2019}, pages 21--27, 2019.

\bibitem{conneau-etal-2020-unsupervised}
Alexis Conneau, Kartikay Khandelwal, Naman Goyal, Vishrav Chaudhary, Guillaume
  Wenzek, Francisco Guzm{\'a}n, Edouard Grave, Myle Ott, Luke Zettlemoyer, and
  Veselin Stoyanov.
\newblock {Unsupervised Cross-lingual Representation Learning at Scale}.
\newblock In {\em Proceedings of ACL}, pages 8440--8451, 2020.

\bibitem{NIPS2019_8928}
Alexis Conneau and Guillaume Lample.
\newblock {Cross-lingual Language Model Pretraining}.
\newblock In {\em NIPS}, pages 7059--7069. 2019.

\bibitem{devlin2019bert}
Jacob Devlin, Ming-Wei Chang, Kenton Lee, and Kristina Toutanova.
\newblock {BERT: Pre-training of Deep Bidirectional Transformers for Language
  Understanding}.
\newblock In {\em Proceedings of NAACL}, pages 4171--4186, 2019.

\bibitem{joulin2017bag}
Armand Joulin, {\'E}douard Grave, Piotr Bojanowski, and Tom{\'a}{\v{s}}
  Mikolov.
\newblock {Bag of Tricks for Efficient Text Classification}.
\newblock In {\em Proceedings of EACL}, pages 427--431, 2017.

\bibitem{kaalep1997estonian}
Heiki-Jaan Kaalep.
\newblock {An Estonian Morphological Analyser and the Impact of a Corpus on its
  Development}.
\newblock {\em Computers and the Humanities}, 31(2):115--133, 1997.

\bibitem{kondratyuk201975}
Dan Kondratyuk and Milan Straka.
\newblock {75 Languages, 1 Model: Parsing Universal Dependencies Universally}.
\newblock In {\em Proceedings of EMNLP-IJCNLP}, pages 2779--2795, 2019.

\bibitem{kuratov2019adaptation}
Yuri Kuratov and Mikhail Arkhipov.
\newblock {Adaptation of Deep Bidirectional Multilingual Transformers for
  Russian Language}.
\newblock {\em arXiv preprint arXiv:1905.07213}, 2019.

\bibitem{liu2019roberta}
Yinhan Liu, Myle Ott, Naman Goyal, Jingfei Du, Mandar Joshi, Danqi Chen, Omer
  Levy, Mike Lewis, Luke Zettlemoyer, and Veselin Stoyanov.
\newblock {Roberta: A Robustly Optimized BERT Pretraining Approach}.
\newblock {\em arXiv preprint arXiv:1907.11692}, 2019.

\bibitem{nozza2020mask}
Debora Nozza, Federico Bianchi, and Dirk Hovy.
\newblock {What the [MASK]? Making Sense of Language-Specific BERT Models}.
\newblock {\em arXiv preprint arXiv:2003.02912}, 2020.

\bibitem{Pajupuu2016IDENTIFYINGPI}
Hille Pajupuu, Rene Altrov, and Jaan Pajupuu.
\newblock {Identifying Polarity in Different Text Types}.
\newblock {\em Folklore}, 64, 2016.

\bibitem{peters2018deep}
Matthew Peters, Mark Neumann, Mohit Iyyer, Matt Gardner, Christopher Clark,
  Kenton Lee, and Luke Zettlemoyer.
\newblock {Deep Contextualized Word Representations}.
\newblock In {\em Proceedings of NAACL}, pages 2227--2237, 2018.

\bibitem{pires2019multilingual}
Telmo Pires, Eva Schlinger, and Dan Garrette.
\newblock {How Multilingual is Multilingual BERT?}
\newblock In {\em Proceedings of ACL}, pages 4996--5001, 2019.

\bibitem{qi2020stanza}
Peng Qi, Yuhao Zhang, Yuhui Zhang, Jason Bolton, and Christopher~D Manning.
\newblock {Stanza: A Python Natural Language Processing Toolkit for Many Human
  Languages}.
\newblock In {\em Proceedings of ACL System Demonstrations}, 2020.

\bibitem{sanh2019distilbert}
Victor Sanh, Lysandre Debut, Julien Chaumond, and Thomas Wolf.
\newblock {DistilBERT, a Distilled Version of BERT: Smaller, Faster, Cheaper
  and Lighter}.
\newblock {\em arXiv preprint arXiv:1910.01108}, 2019.

\bibitem{straka2017tokenizing}
Milan Straka and Jana Strakov{\'a}.
\newblock {Tokenizing, POS Tagging, Lemmatizing and Parsing UD 2.0 with
  UDPipe}.
\newblock In {\em Proceedings of the CoNLL 2017 Shared Task}, pages 88--99,
  2017.

\bibitem{tkachenko2013named}
Alexander Tkachenko, Timo Petmanson, and Sven Laur.
\newblock Named entity recognition in {E}stonian.
\newblock In {\em Proceedings of the 4th Biennial International Workshop on
  {B}alto-{S}lavic Natural Language Processing}, pages 78--83, 2013.

\bibitem{virtanen2019multilingual}
Antti Virtanen, Jenna Kanerva, Rami Ilo, Jouni Luoma, Juhani Luotolahti, Tapio
  Salakoski, Filip Ginter, and Sampo Pyysalo.
\newblock {Multilingual is not enough: BERT for Finnish}.
\newblock {\em arXiv preprint arXiv:1912.07076}, 2019.

\end{thebibliography}

\end{document}